\documentclass[conference]{IEEEtran}
\usepackage{cite}
\usepackage{amsmath,amssymb,amsfonts}
\usepackage{algorithmic}
\usepackage{graphicx}
\usepackage{textcomp}
\usepackage{xcolor}
\usepackage{url}
\def\BibTeX{{\rm B\kern-.05em{\sc i\kern-.025em b}\kern-.08em
    T\kern-.1667em\lower.7ex\hbox{E}\kern-.125emX}}
\begin{document}

\title{Concept Drift Detection in Federated Networked Systems}
\author{Dimitrios~Michael~Manias, Ibrahim~Shaer, Li~Yang, and Abdallah~Shami\\
ECE Department, Western University, London ON, Canada\\
\{dmanias3, ishaer, lyang339, Abdallah.shami\}@uwo.ca}
\maketitle
\begin{abstract}
As next-generation networks materialize, increasing levels of intelligence are required. Federated Learning has been identified as a key enabling technology of intelligent and distributed networks; however, it is prone to concept drift as with any machine learning application. Concept drift directly affects the model's performance and can result in severe consequences considering the critical and emergency services provided by modern networks. To mitigate the adverse effects of drift, this paper proposes a concept drift detection system leveraging the federated learning updates provided at each iteration of the federated training process. Using dimensionality reduction and clustering techniques, a framework that isolates the system's drifted nodes is presented through experiments using an Intelligent Transportation System as a use case. The presented work demonstrates that the proposed framework is able to detect drifted nodes in a variety of non-iid scenarios at different stages of drift and different levels of system exposure.
\end{abstract}


\section{Introduction}

With the rapid development of next-generation networking systems and technologies, the use of Machine Learning (ML) and Artificial Intelligence (AI) has cemented itself in the future of networking. As the 5th generation and beyond networks (5G+) take shape, their reliance on intelligence is profound; with requirements such as self-healing, self-configuration and forecasting, ML and AI have revolutionized modern networks and networking practices \cite{A1}. However, the application of such technologies in the realm of networking is still in its infancy. Despite its various benefits, several challenges are still yet to be addressed. One of the most significant challenges faced by ML models is the idea of \textit{model drift}, where changes impacting the domain of a model affect its performance. Some examples of drift in networks include traffic changes caused by crowd events and new movie releases in CDNs. When considering the types of applications hosted by network systems and preserving Quality of Service (QoS) guarantees, it is critical to implement mechanisms to detect the deterioration of the ML model and mitigate the situation before severe consequences are faced. When considering use cases such as Intelligent Transportation Systems (ITSs), consisting of vehicular clients and pedestrians, the consequences can be deadly. \par

One type of model drift is concept drift, which describes a situation where the underlying relationship between inputs and outputs has changed. This type of drift is challenging to diagnose, and it can manifest itself gradually over time \cite{RV1}. While many methods currently exist to detect concept drift in ML-enabled systems, many of them require excessive storage and processing capabilities. When considering emerging technologies such as Multi-access Edge Computing (MEC), which pushes computational power to the edge of the network through the use of lightweight points of presence, these methods are rendered infeasible. As such, it is critical to develop lightweight, scalable, and efficient drift detection techniques for highly distributed and networked systems \cite{A18}. \par

One of the great advancements of ML techniques in recent times has been the introduction of Federated Learning (FL). FL is a decentralized and distributed ML technique that is composed of several federated nodes. Each of these nodes collects, stores, and processes its own information. Due to its ability to preserve local node data privacy, FL has been identified as an enabling technology for next-generation networking systems, and specifically, its use in ITSs has been gaining significant traction \cite{A4}. Since FL relies on lightweight distributed points of presence, it is a prime candidate for developing a concept drift detection system capable of addressing the limitations of current methods. \par

The premise behind FL is that an entity known as the aggregation agent initializes and distributes a global model to each federated node. Using their locally collected data, these nodes perform training iterations, thereby creating a local model. After a predefined number of iterations, the local model is compared to the initially distributed global model, and the differences between the two are formatted as an update. Each node forwards its update to the aggregation agent, which collects and aggregates them to develop a new global model. This new model is redistributed to the federated nodes, and the process repeats. Fig. 1 outlines the federated learning process.

\begin{figure}[!htbp]
\centerline{\includegraphics[width=0.9\columnwidth]{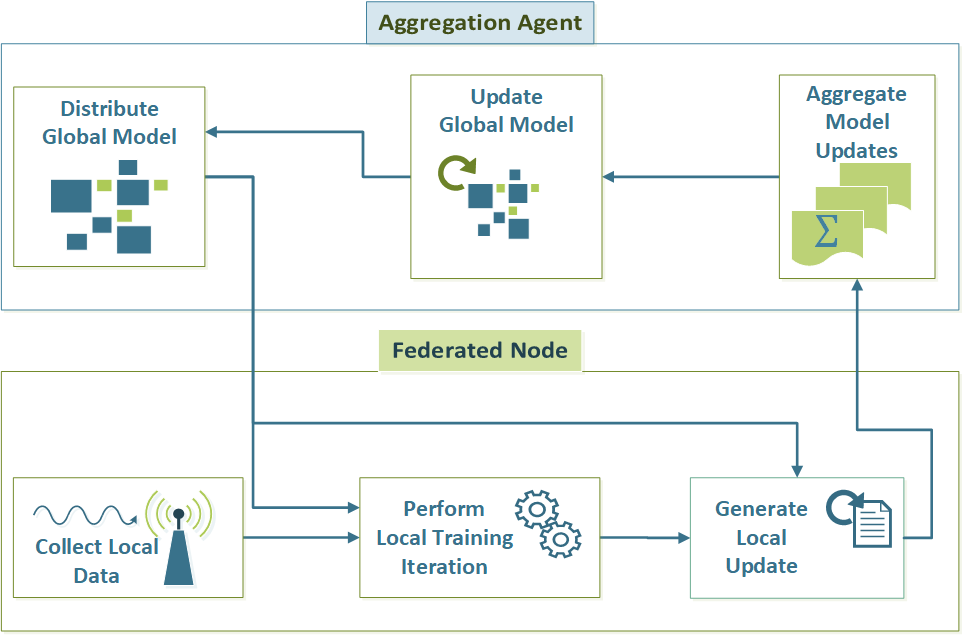}}
\caption{Federated Learning Overview}
\end{figure}

An ITS is the culmination of various interacting networking technologies and architectures, including Network Function Virtualization (NFV), MEC, 5G+ networks, the Internet of Vehicles (IoV) as well as ML and AI. Through distributed points of presence known as Roadside Units (RSUs), an ITS is able to collect and process data at the edge of the network and provide functionalities to vehicular clients \cite{RV2}. The basic setup of an ITS is presented in Fig. 2. 

\begin{figure}[!htbp]
\centerline{\includegraphics[width=0.9\columnwidth]{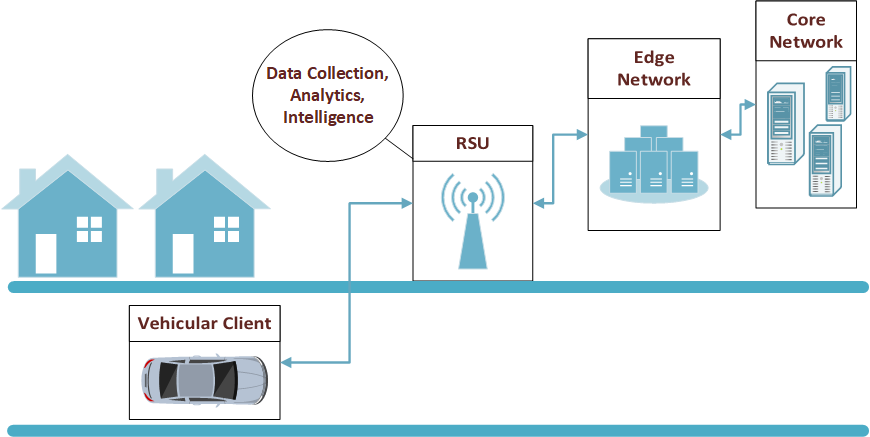}}
\caption{Basic ITS Overview}
\end{figure}

The ITS has been selected as a use case for the work presented in this paper due to its use of various networking technologies as well as its establishment as a critical future networking architecture; however, the proposed framework is applicable to any distributed networked system. \par

In this work, the distributed and collaborative nature of FL is leveraged to identify and isolate nodes that are experiencing concept drift. By observing and modelling the behaviour of the federated training process under normal (absence of drifted concepts) conditions, any future training iteration can be compared to the expected behaviour. If a significant statistical difference between the most recent federated training iteration updates and the normal modelled updates is observed, a concept drift alarm is raised, and the nodes exhibiting the drift are identified. By identifying drifted nodes, isolation measures can be enacted to ensure that the remainder of the system, still operating under normal circumstances, does not experience performance degradation as a result of the drift.  \par

While several proposed concept drift detection systems exist, several limitations prevent their implementation in networked systems. The work outlined in this paper proposes a lightweight and distributed concept drift detection framework through the use of FL. By leveraging various updates provided to the aggregation agent by each federated node, this framework allows for a resource-efficient method of detecting concept drift in a highly dynamic system environment. The contributions of this paper are summarized as follows:
\begin{itemize}
\item The development of a federated framework leveraging principal component analysis and K-means clustering to isolate drifted nodes in a distributed networked system.
\item The presentation of a use case highlighting the applicability of a federated framework for concept drift detection in an ITS.
\item The analysis of the system through various drift scenarios.
\end{itemize} \par

The remainder of this paper is structured as follows. Section II outlines the related work regarding concept drift detection and FL in ITSs. Section III discusses the proposed methodology. Section IV outlines the implementation. Section V presents the results and analysis. Finally, Section VI concludes the paper and presents opportunities for future work.

\section{Related Work}

The following is a list of concept drift detection and mitigation techniques for traditional machine learning systems. Gama \textit{et al.} \cite{A5} monitor an online error rate and use a threshold value to determine if a concept has drifted. Garcia \textit{et al.} \cite{A6} use the distribution of distances between errors to detect concept drift; this method specifically targets gradual drifts over time as opposed to sudden drifts. Sun \textit{et al.} \cite{A7} propose combining past and current models as a method of incremental learning to combat drift. Widmer and Kubat \cite{A8} use a sliding window and develop a history of previously encountered concepts that are expected to reappear in a future period. Elwell and Polikar \cite{A9} suggest a weighted ensemble learner which trains a new model with each incoming batch of data. Forman \cite{A10} discusses the implementation of an ensemble where a model is trained daily on incoming data and the predictions of previous models.  \par

While several methods addressing concept drift detection and mitigation exist, they are inadequate for implementation in FL systems due to their reliance on previous data and models. One of the key advantages of federated systems is that they are decentralized and distributed; therefore, the storage of data and models at a central location is infeasible. Additionally, since the federated networked systems deal with lightweight points of presence at the network edge, the excessive use of storage is an inefficient utilization of limited resources.

The application of FL to ITSs has gained significant traction in recent years. Manias and Shami discuss the need for FL as a method of advanced intelligence to be applied to the management and orchestration of NFV-enabled networks \cite{A11}, and they specifically make a case for using FL in an ITS \cite{A4}. Elbir \textit{et al.} \cite{A12} discuss the benefits and challenges of FL applied to vehicular networks. Khan \textit{et al.} \cite{A13} suggest FL for resource optimization in edge networks and mention its applicability for distributed 5G-enabled applications, including ITSs. Finally, Du \textit{et al.} \cite{A14} discuss how FL can address challenges such as a large number of connected devices and the preservation of privacy in the Vehicular IoT and ITSs. As demonstrated by the work above, FL is actively being discussed and implemented in next-generation networking systems and use cases. To ensure its feasibility, an effort to develop lightweight and effective concept drift detection and mitigation schemes capable of addressing the limitations of current methods applied to a resource-constrained MEC-enabled environment is critical.

\section{Methodology}

The following section will discuss the methodology presented in this work.

\subsection{Framework Architecture}
The proposed federated concept drift detection framework consists of two phases, the system training phase and the active drift detection phase. During the system training phase, it is assumed that no concept drift is present; as such, the observations collected during this phase form the basis of what is categorized as normal behaviour. In this stage, the FL process proceeds as normal. The federated nodes collect local data, perform local training iterations, and develop a model update. At this point, the update is sent to the aggregation agent as part of the FL process; however, it is also sent to the drift detection module hosted on the node. This drift detection module is composed of three stages, Principal Component Analysis (PCA), K-Means clustering, and distance calculation.  \par
The role of PCA in the drift detection module is to reduce the dimensionality of the weight portion of the model update to improve both the storage and communication efficiency of the module as well as to eliminate the correlation between specific weight updates. The optimal number of principal components as part of PCA is determined by locating the point of inflection in the plot of explained variance per component. The reduced weight updates are sent to the normal weight update storage until the system training process completes. \par

After a predefined number of training iterations across the system are completed, the drift detection module applies K-Means clustering on the reduced weight updates and creates two clusters. K-Means was selected due to its speed and efficiency as well as having a priori knowledge of the required number of clusters; Additional clustering methods, including hierarchical clustering, will be explored in future work. The resulting cluster centers are then computed and passed to the distance calculator, which computes the Euclidean distance between the two cluster centers. Depending on the optimal number of principal components selected, the Euclidean distance is calculated across multiple dimensions according to Eq. 1 where $dist(c_{1}, c_{2})$ denotes the distance between cluster centers $c_{1}$ and $c_{2}$, and $n$ is the number of principal components (dimensions).

\begin{equation}
dist(c_{1},c_{2}) = \sqrt{\sum_{i}^{n}(c_{2_{i}}-c_{1_{i}})^2}
\end{equation}

At this point, each node in the system, having completed its system training and calculating the cluster distances for its normal operation, sends the resulting distance to the aggregation agent to be stored in the normal model statistics. In this module, the aggregation agent considers the cluster distance across all nodes and calculates the mean and standard deviation of the cluster distances, which will be used to determine if a node has drifted or not during the drift detection stage through thresholding. This concludes the system training phase, and the framework is now ready for active drift detection. \par

During the active drift detection phase, a similar process is followed. Firstly, the normal model statistics are retrieved from the aggregation agent and passed to the drift detection module. This happens once, at the beginning of the process, as these statistics aren’t updated further. When a model update is generated, it is passed to the drift detection module along with the contents of the normal weight update storage, which, as previously mentioned, contains the reduced weights obtained during the system training phase. Once in the drift detection module, its weights are reduced and added to the active drift detection storage. At a specified detection interval (\textit{i.e.} after $n$ active drift detection iterations), clustering is performed on the combination of the active drift detection reduced weights along with the normal reduced weights, and the cluster distance is calculated. An interval is selected as not constantly to burden the node with unnecessary computation. The newly calculated distance is compared to the normal model statistics and thresholding is applied to determine if a drift has occurred. Under this thresholding, a node is considered to be drifted if the distance between its cluster centers is more than 3 standard deviations away from the normal training mean. This threshold was determined by examining the performance of the system and will be explored further in future work.  This threshold is formalized in Eq. 2, where $\mu_{norm}$ denotes the mean cluster distance observed during the normal training process, $\sigma_{norm}$ denotes the standard deviation observed during the normal training process, and $c_{dist}$ denotes the observed cluster distance during the drift detection phase. 

\begin{equation}
c_{dist} < \mu_{norm} - 3\sigma_{norm} \lor \mu_{norm} + 3\sigma_{norm} < c_{dist} 
\end{equation}

If a drift is detected, the node sends a drift warning to the aggregation agent; otherwise, it sends the model update. An overview of the system training and active drift detection phases is presented in Fig. 3, where the interactions between the federated nodes and the aggregation agent are displayed. Moreover, Fig. 4 delves deeper into the active drift detection phase and presents a granular process map. Additionally, this figure highlights how the drift detection process presented is added to the local federated learning process. The interface between a federated node and the aggregation agent is depicted through a dashed line.

\begin{figure}[!htbp]
\centerline{\includegraphics[width=0.9\columnwidth]{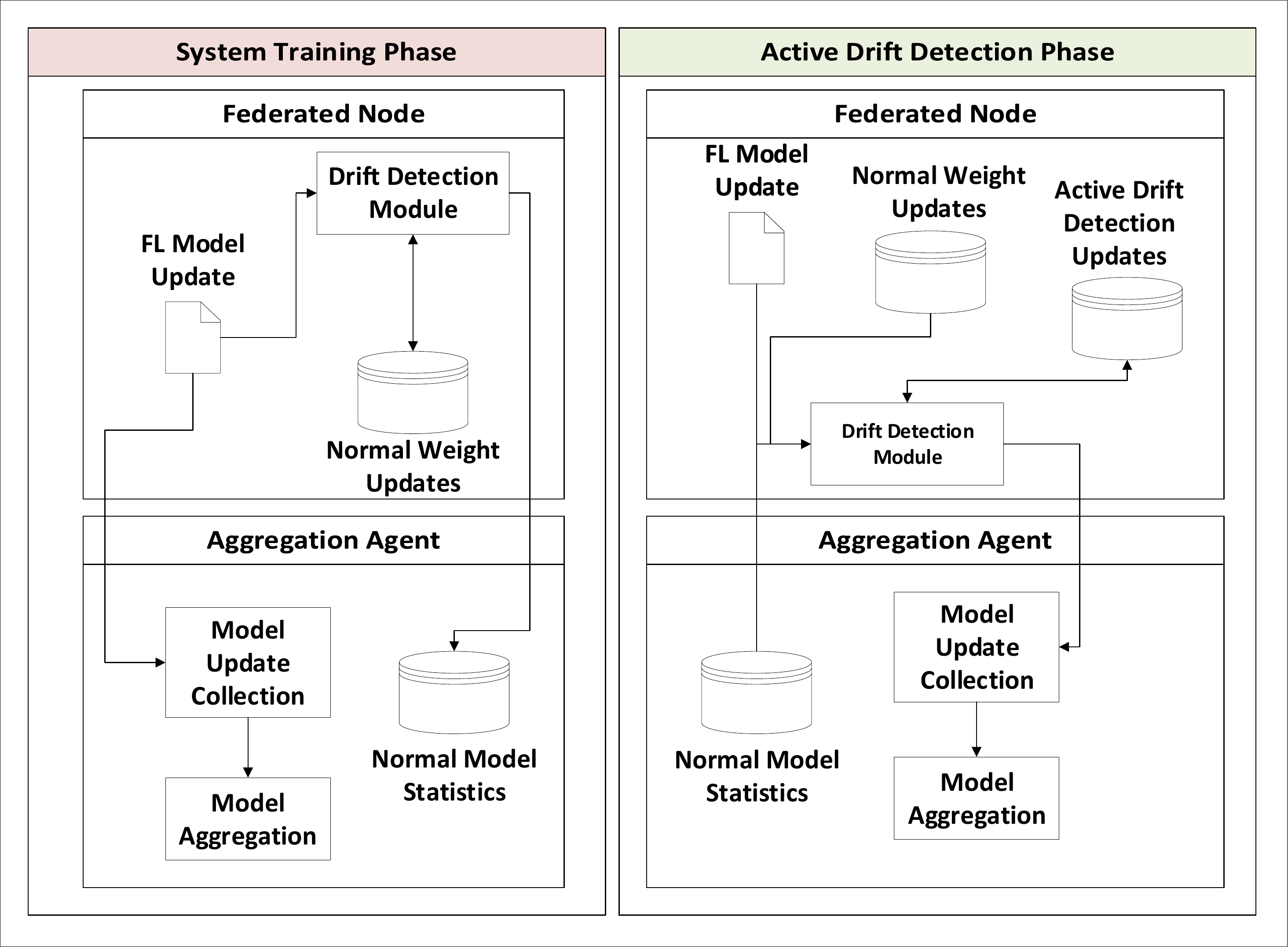}}
\caption{System Training vs. Active Drift Detection Phases}
\end{figure}

\begin{figure}[!htbp]
\centerline{\includegraphics[width=0.9\columnwidth]{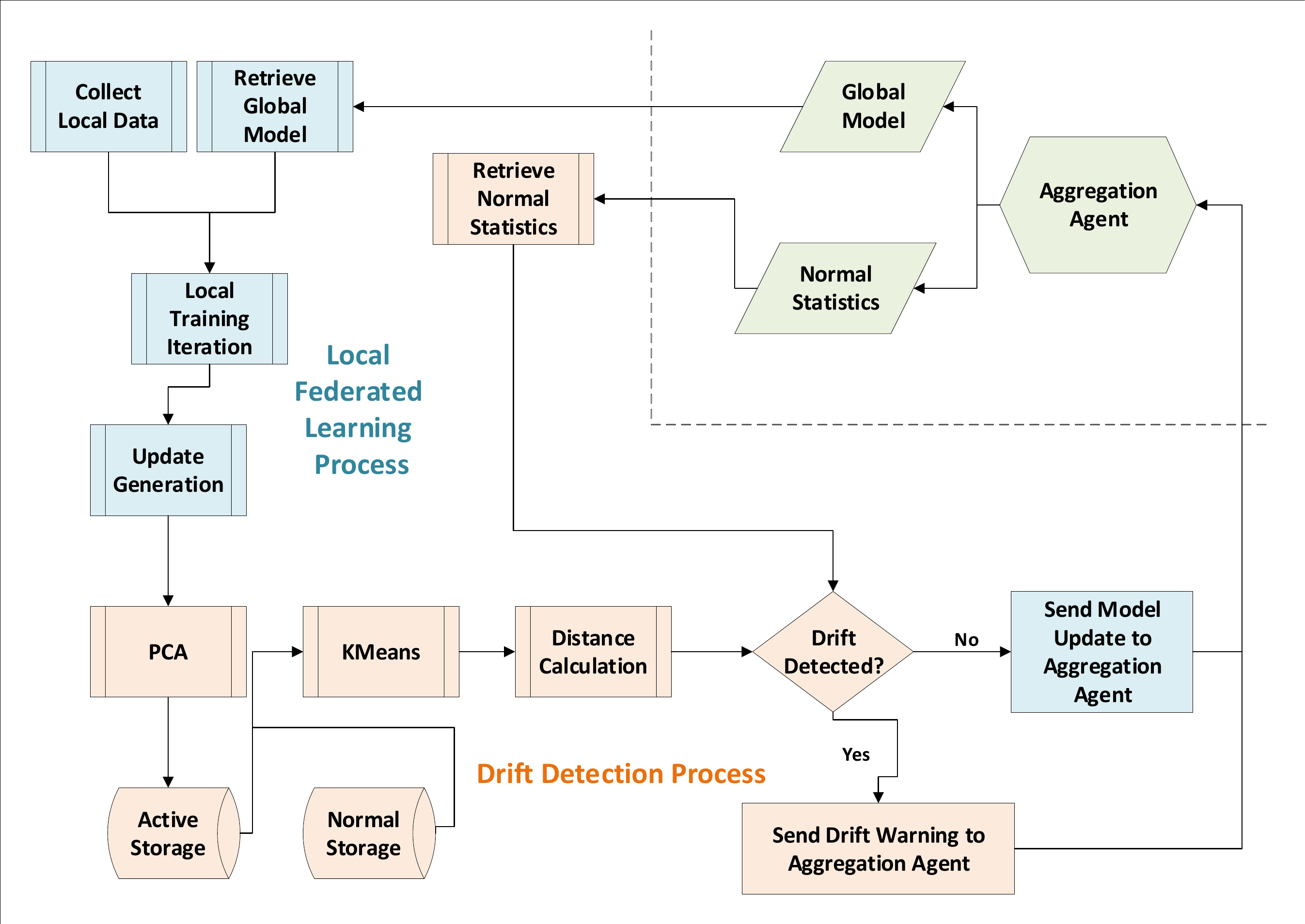}}
\caption{Active Drift Detection Phase Process Map}
\end{figure}

\subsection{System Architecture}
As previously mentioned, the proposed concept drift detection framework can be applied to any networked federated system. Specifically, in this work, an ITS is considered. The premise of this system is that a group of geo-distributed RSUs are collecting image data of the vehicular clients they encounter and are performing a vehicular traffic classification task. Each of the RSUs uses a neural network model to perform the classification and is part of a system leveraging FL. At some point in time, an event causing a reclassification of certain vehicular clients occurs (\textit{e.g.} new legislations or the repurposing of once public vehicles for private use) which causes a fundamental drift in the system. While the effect of this drift is sudden, it propagates through the system by affecting a subset of the federated nodes. The deployed drift detection framework actively works to locate drifted nodes and allow the aggregation agent to enact isolation measures.

\subsection{System Communication Metrics}

There are several metrics that have been developed to assess the communication utilization of this drift detection method. Consideration of these metrics is motivated by the highly distributed and resource-constrained nature of this system. The first metric considered is the system training phase communication ratio, $\alpha$, which determines the impact of the results of the normal cluster distances being communicated along with the regular updates as part of the FL process. For this calculation, the number of training iterations is denoted by $n_{iter}$, the size of the model update is denoted by $l_{update}$, and the size of the distance information is denoted by $l_{dist}$. The formulation of the ratio is presented in Eq. 3. 

\begin{equation}
\alpha = \frac{(n_{iter} \times l_{update})+l_{dist}}{n_{iter} \times l_{update}}
\end{equation}

Considering that the size of the cluster distance is negligible compared to the model update, the $\alpha$ ratio demonstrates that the additional information being sent to the aggregation agent at the end of the system training phase has little impact on the communication resources being consumed. \par

The second metric considered is the communication ratio during the active drift detection phase, $\beta$ which determines the quantity of resources being conserved by not sending drifted node updates to the aggregation agent. In this ratio, $n$ denotes the number of system nodes, $d$ denotes the number of drifted nodes, and $l_{alert}$ denotes the size of the drift alert message. This ratio is presented in Eq. 4. 

\begin{equation}
\beta = 1 - \frac{(n-d) \times l_{update} + d \times l_{alert}}{n \times l_{update}}
\end{equation}

A limit analysis, $\lim_{d \to n} \beta \approx 1$ highlights how the amount of communication resources conserved increases as the number of drifted nodes increases. Considering that the size of the update is orders of magnitude greater than the size of the alert, the amount of communication resources saved by halting drifted nodes from sending their updates is evident.

\section{Implementation}
The implemented system consists of 10 RSUs, each acting as a federated node capturing images of vehicular traffic. The system training phase consists of 100 iterations.  The federated learning process uses the federated averaging aggregation scheme as discussed by McMahan \textit{et al.} \cite{A15}. The MNIST digit dataset \cite{A16} is used to simulate types of vehicular traffic; this dataset was selected for its widespread use for benchmarking image classification tasks. This dataset contains 60000 samples of handwritten digits ranging from 0 to 9. For simulation purposes, each digit represents a type of vehicular traffic (\textit{i.e.}, sedan, SUV, semi, bus, \textit{etc.}). Initially, digits 0 through 5 are assigned to class 1, and digits 6 through 9 are assigned to class 2. The two classes can represent a myriad of possible labels in ITS vehicular traffic classification tasks (\textit{e.g.} industrial vs. residential, normal vs. abnormal, private vs. public, \textit{etc.}). During a concept drift scenario, the fundamental relationship between the digits and their assigned class is altered. To simulate this, a digit is randomly selected from class 2 and is converted to class 1. This change in the underlying relationship between the digits and their associated classes illustrates a fundamental concept drift and therefore forms the basis of our experiments. \par
The model used to perform the image classification task is a shallow artificial neural network. As the images provided in the dataset are 28 pixels x 28 pixels, they form an array of 784 input features once flattened. The output layer consists of two neurons, one for each of the possible binary classes. A softmax activation function is used to determine the classification decision. The implementation of the model was done in Python using the TensorFlow and TensorFlow Federated packages. The application of the method proposed in this work on deeper neural network architectures is out of the scope of this paper and will be explored in future work. 

\subsection{Experiments}

When FL was introduced in 2016 \cite{A17}, one of the key assumptions was that the data across all nodes was non-IID, which was a paradigm shift from traditional ML techniques. As such, all experiments presented operate on the assumption that the data are non-IID. The first experiment conducted as part of this work considers an extreme scenario where each of the ten nodes has access to data of a single digit. \par

In this scenario, each of the federated nodes contains 1000 training examples. The system training phase proceeds for 100 iterations. In these experiments, the dimensionality of the normal training weight updates is reduced by more than 99.7\%, greatly reducing the amount of data stored as part of the proposed drift detection process after using PCA.  \par

The drift is then simulated by distributing 1000 new and previously unseen examples to each node and changing the class associated with one of the digits belonging to class 2 as previously defined. This marks the start of the drift detection phase, which proceeds with an interval of 10 updates. Finally, a drift decision is made after clustering and thresholding. \par

The implementation for the remaining experiments is similar to what has been described; however, the nodes do not have access to only a single digit. This means that each node has differing amounts of data and different distributions across observations. To simulate the non-IID data distributions, the number of samples for each digit is randomly selected from the range [100,500]. This means that the minimum number of observations a node can have is 1000, and the maximum is 5000. This is consistent with the simulation of an ITS environment as certain RSUs are located in low-traffic areas, whereas others are located in high-traffic areas. Additionally, in this scenario, single and multiple node drifts are evaluated. In the second experiment, a single node experiences the drift; in the third experiment, two nodes experience the drift, and in the final experiment, four nodes experience the drift. These experiments highlight the efficacy of the proposed solution across various levels of system drift. \par

\section{Results and Analysis}

The following section presents and analyzes the results obtained from each of the experiments.

\subsection{Experiment 1}

The first part of analyzing the results from experiment 1 was to determine the optimal number of principal components for PCA. As previously mentioned, this was done by plotting the explained variance across the first ten principal components yielding an elbow plot indicating the optimal number of principal components as two. After applying PCA on each node's respective updates, K-Means clustering was applied. Since this experiment resulted in one drifted node, both the effects of clustering of drifted and un-drifted nodes must be examined. Figure 5 presents the clustering of the reduced weight updates for both a drifted and un-drifted node.

\begin{figure}[!htbp]
\centerline{\includegraphics[width=\columnwidth]{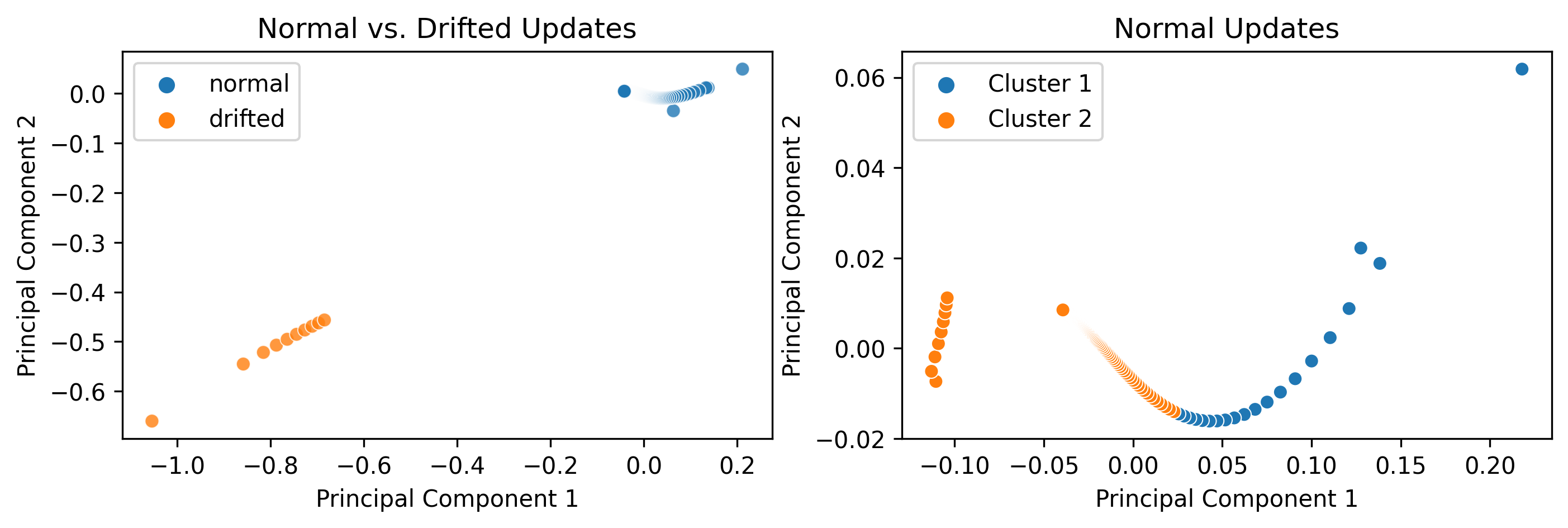}}
\caption{Drifted Node Update Clustering}
\end{figure}

As seen in the left graph of Fig. 5, there is a clear distinction between the drifted updates and the normal updates. Comparatively, in the right graph, the results of the clustering for the normal node are not as decisive as those observed in the drifted node due to the proximity of the two clusters. This observation is what led to the next step of the methodology to determine the distance between the two cluster centers as a method of identifying the drifted nodes. Fig. 6 presents the distance between the two center clusters of each node. In this figure and all subsequent figures, the threshold boundaries $\mu_{norm} \pm 3\sigma_{norm}$ are depicted as the blue (mean) and cyan (standard deviation) dashed  lines, respectively. 

\begin{figure}[!htbp]
\centerline{\includegraphics[width=0.875\columnwidth]{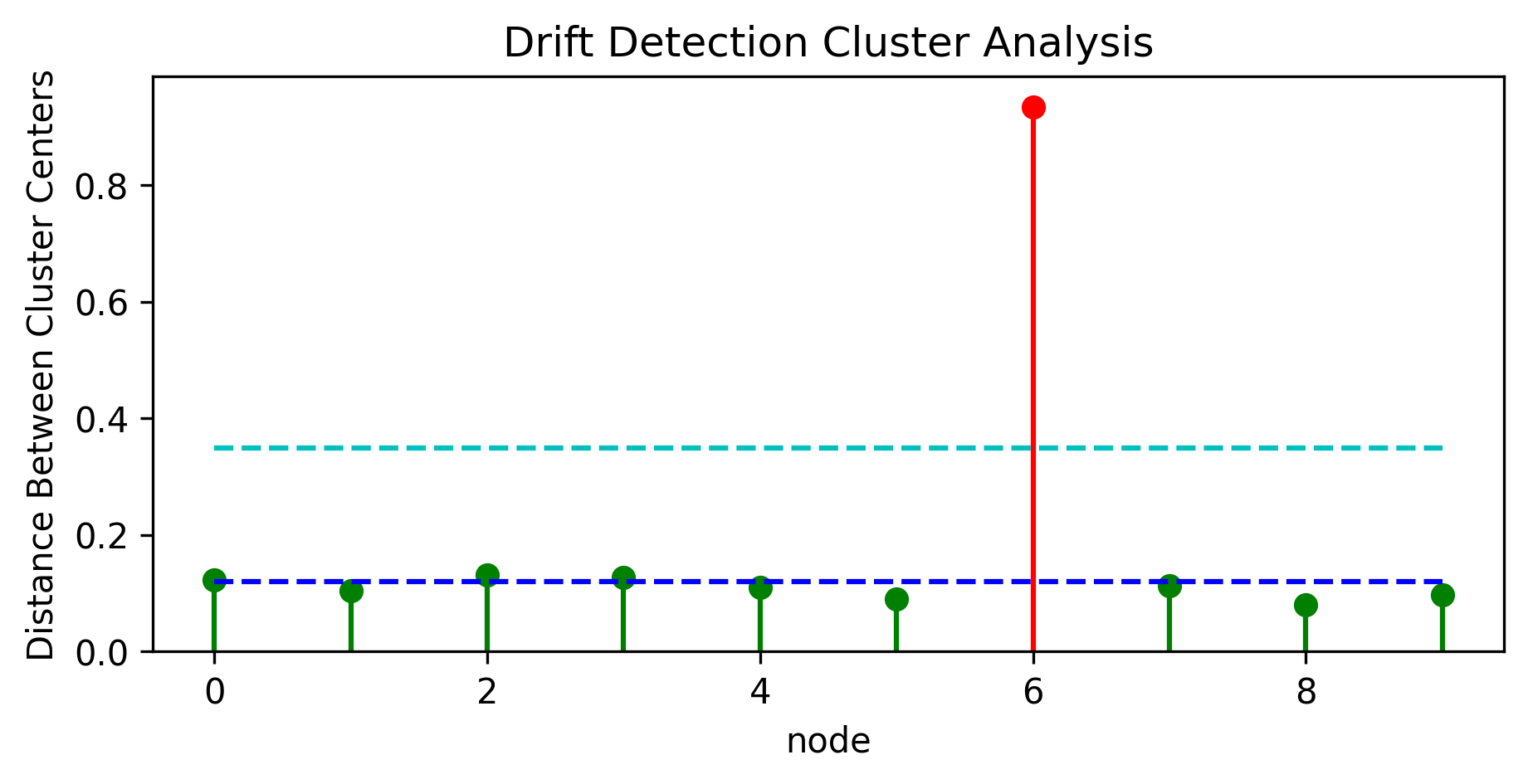}}
\caption{Experiment 1 Drift Detection}
\end{figure}

As seen through Fig. 6, node six, labelled in red, has been correctly identified as the drifted node as its cluster center distance is significantly different from what is observed across the remaining nodes. It should be noted that since one node has drifted, the $\beta$ metric is approximately equal to 0.1, indicating a 10\% reduction in wasteful communication.

\subsection{Experiment 2}

In experiment 2, the point of inflection in the elbow graph corresponds to four principal components. Regarding the clustering and distance calculation, the proposed framework correctly identified node seven as the drifted node. However, it should be noted that in this scenario, the distance between drifted and non-drifted nodes is less than what was observed in Fig. 6 of experiment 1. This can be attributed to the fact that this experiment (and all subsequent experiments) has a more complex system setup as each node has access to data of all digits rather than being exclusive to one. Results of this cluster analysis and the identification of the drifted node, seven, are presented in Fig. 7. Similarly to experiment 1, the $\beta$ ratio for this experiment was also 0.1.

\begin{figure}[!htbp]
\centerline{\includegraphics[width=0.875\columnwidth]{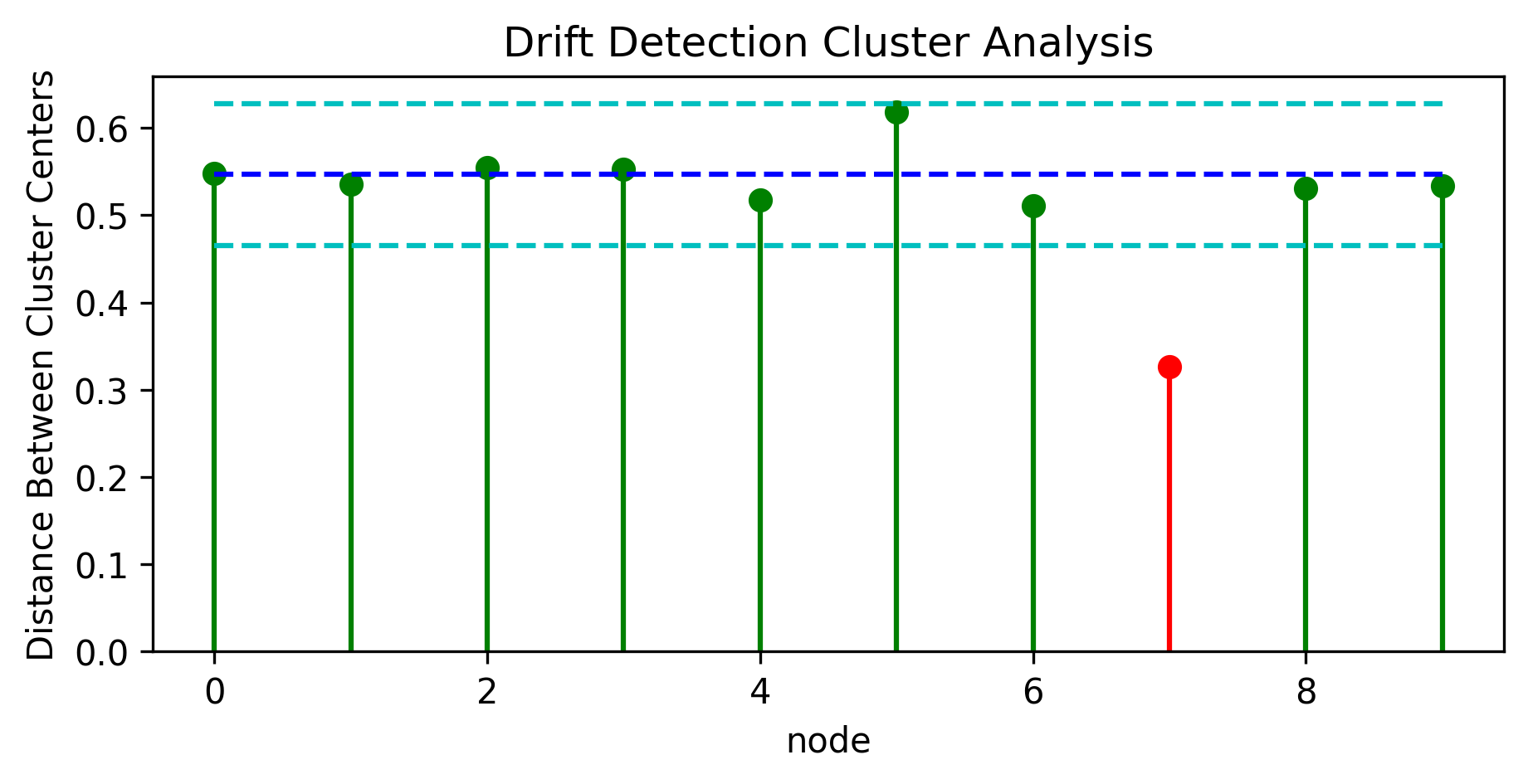}}
\caption{Experiment 2 Drift Detection}
\end{figure}

\subsection{Experiment 3}
The number of principal components used in this experiment was also 4. The cluster analysis results successfully yielded the two drifted nodes, six and nine, as seen in Fig. 8. Since there are two drifted nodes, the $\beta$ ratio for this experiment is approximately 0.2.

\begin{figure}[!htbp]
\centerline{\includegraphics[width=0.875\columnwidth]{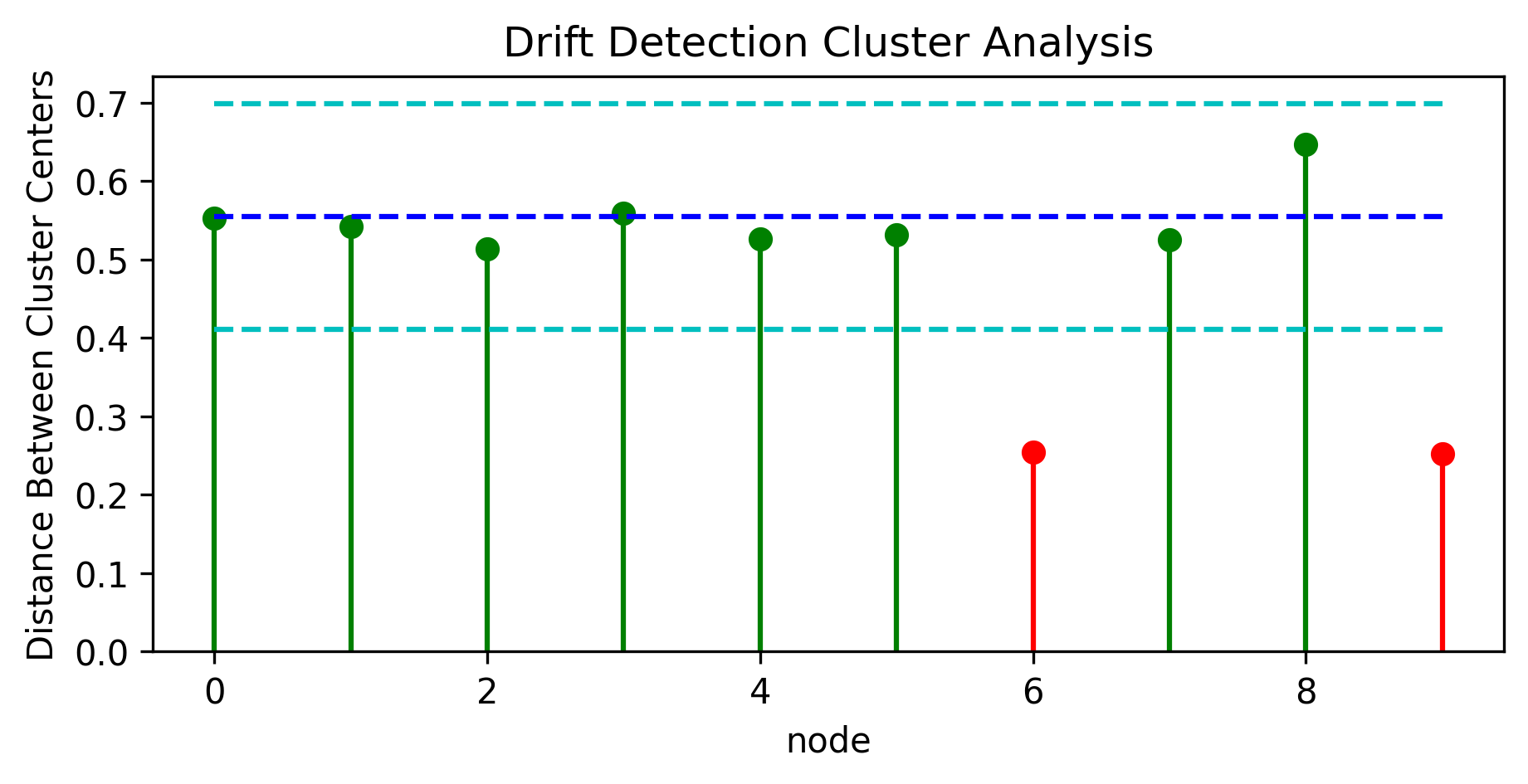}}
\caption{Experiment 3 Drift Detection}
\end{figure}

\subsection{Experiment 4}
Similar to experiments 2 and 3, four principal components were used in experiment 4. The cluster analysis results successfully yielded the drifted nodes, which were 1, 3, 4, 5, as seen in Fig. 9. Additionally, since there are four drifted nodes, the $\beta$ ratio for this experiment is approximately 0.4.

\begin{figure}[!htbp]
\centerline{\includegraphics[width=0.875\columnwidth]{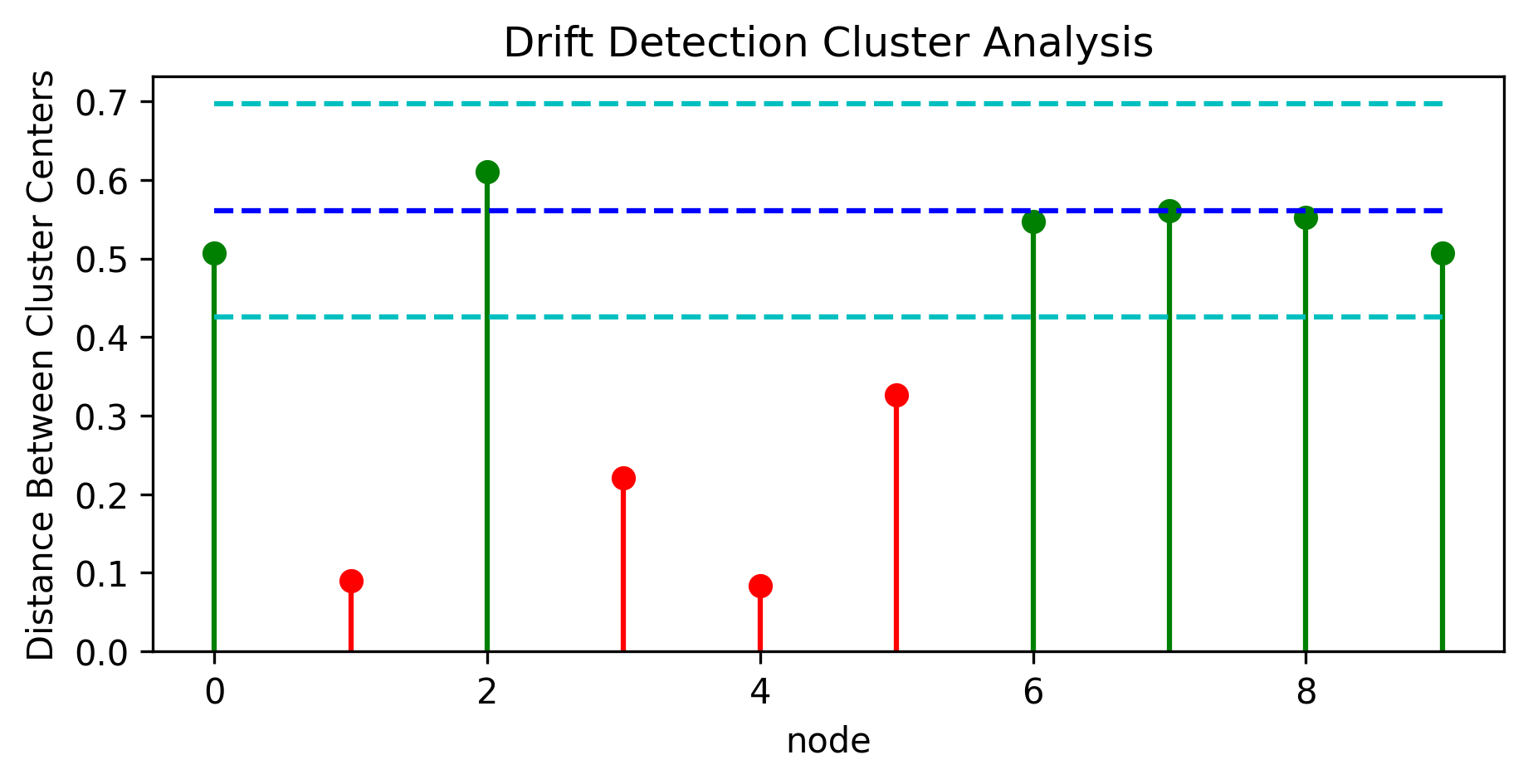}}
\caption{Experiment 4 Drift Detection}
\end{figure}

\section{Conclusion}
In conclusion, the work presented in this paper proposes a framework for detecting concept drift in federated networked systems by leveraging the updates sent by the federated nodes during each iteration of the FL process. By using PCA to reduce the dimensionality of the weight updates, applying K-Means clustering, and calculating the distance between cluster centers, drifted nodes are identified. The work presented in this paper highlights ITSs as a use case; however, it is applicable to any federated networked system. \par

Future work will explore concept drift detection frameworks for different ML tasks such as multi-class classification as well as regression tasks. Additionally, deeper network architectures will be explored to evaluate the optimal level of dimensionality reduction for more complicated models. Furthermore, different thresholding schemes to isolate drifted nodes will be explored. Finally, larger network sizes with an increasing number of federated nodes will be assessed for scalability.

\end{document}